\documentclass[10pt,twocolumn,letterpaper]{article}

\usepackage[pagenumbers]{cvpr} 

\usepackage{graphicx}
\usepackage{amsmath}
\usepackage{amssymb}
\usepackage{booktabs}
\usepackage{enumitem}
\usepackage{caption}
\usepackage{multirow}
\usepackage{makecell}
\usepackage{floatrow}
\usepackage{flushend}

\usepackage[flushleft]{threeparttable}

\usepackage[pagebackref,breaklinks,colorlinks]{hyperref}
\usepackage{utfsym} 

\usepackage[capitalize]{cleveref}
\crefname{section}{Sec.}{Secs.}
\Crefname{section}{Section}{Sections}
\Crefname{table}{Table}{Tables}
\crefname{table}{Tab.}{Tabs.}

\hypersetup{
        colorlinks=true,
        linkcolor=blue,
        filecolor=green,      
        urlcolor=blue, 
        citecolor=red,
}

\begin{document}

\title{SHapley Estimated Explanation (SHEP): A Fast Post-Hoc Attribution Method for Interpreting Intelligent Fault Diagnosis}

\author{Qian~Chen, Xingjian~Dong, Zhike~Peng, Guang~Meng\\
\textit{State Key Laboratory of Mechanical System and Vibration, Shanghai Jiao Tong University}\\
{\tt\small chenqian2020@sjtu.edu.cn}
}

\maketitle

\begin{abstract}

  Despite significant progress in intelligent fault diagnosis (IFD), the lack of interpretability remains a critical barrier to practical industrial applications, driving the growth of interpretability research in IFD. Post-hoc interpretability has gained popularity due to its ability to preserve network flexibility and scalability without modifying model structures. However, these methods often yield suboptimal time-domain explanations. Recently, combining domain transform with SHAP has improved interpretability by extending explanations to more informative domains. Nonetheless, the computational expense of SHAP, exacerbated by increased dimensions from domain transforms, remains a major challenge.
  To address this, we propose patch-wise attribution and SHapley Estimated Explanation (SHEP). Patch-wise attribution reduces feature dimensions at the cost of explanation granularity, while SHEP simplifies subset enumeration to approximate SHAP, reducing complexity from exponential to linear. Together, these methods significantly enhance SHAP's computational efficiency, providing feasibility for real-time interpretation in monitoring tasks. Extensive experiments confirm SHEP's efficiency, interpretability, and reliability in approximating SHAP. Additionally, with open-source code, SHEP has the potential to serve as a benchmark for post-hoc interpretability in IFD.
  The code is available on \textit{\url{https://github.com/ChenQian0618/SHEP}}.
\end{abstract}

\section{Introduction}
Mechanical fault diagnosis (FD), as a key aspect of prognostics and health management (PHM), identifies fault types or locations to facilitate targeted maintenance. This effectively reduces operational costs and repair time, making it widely applied in industrial production and equipment maintenance~\cite{leiMachineryhealthprognostics2018}.

Traditional FD research is based on dynamics~\cite{dongIdentificationerrorexcitation2024} and signal processing techniques~\cite{gangsarSignalbasedcondition2020,zhangVibrationfeatureextraction2022,fengreviewvibrationbasedgear2023}, employing targeted modeling and analysis of fault mechanisms and signal characteristics~\cite{randallVibrationBasedConditionMonitoring2021}. While these methods offer a solid theoretical foundation and strong interpretability, they are time-consuming and labor-intensive due to their reliance on prior knowledge. Advances in sensor technology and modern machinery
have led to widespread sensor deployment, generating vast amounts of operational data~\cite{leiApplicationsmachinelearning2020}. This transition to the big data era poses significant challenges for traditional FD method. Conversely, data-driven intelligent fault diagnosis (IFD) offers high efficiency, accuracy, and end-to-end solutions without relying on prior knowledge, making it a promising alternative for fault diagnosis in the big data era~\cite{qianFederatedtransferlearning2025}.

IFD utilizes neural networks to learn the nonlinear mapping between operational signals (typically vibration signals) and equipment fault states. It has emerged as a thriving research field, with advanced networks increasingly applied to diagnosing mechanical components such as bearings, gears, and pumps~\cite{lvAttentionmechanismintelligent2022}. Despite significant advancements in diagnostic accuracy~\cite{michauFullylearnabledeep2022}, generalization~\cite{yangLabelRecoveryTrajectory2024}, information fusion~\cite{sunMultisensortemporalspatialgraph2025}, and few-shot learning~\cite{shaoFewShotCrossDomainFault2024}, the interpretability of IFD remains underexplored~\cite{zhaoChallengesOpportunitiesAIEnabled2021}.

Neural network, as the foundation of IFD, involves multiple nonlinear mappings and is a typical ``\textit{black-box}"~\cite{zhangVisualinterpretabilitydeep2018}. This lack of interpretability makes it difficult to understand their reasoning, inference logic, and scope of application. Nevertheless, interpretability is vital for the practical implementation of IFD~\cite{ivanovsPerturbationbasedmethodsexplaining2021}.
From the user's perspective, an uninterpretable IFD struggles to gain trust, necessitating additional cross-validation and incurring higher costs.
From the developer's perspective, the lack of interpretability hinders systematic error correction and optimization, forcing reliance on empirical or trial-and-error methods.
From the application perspective, real-world scenarios are more complex and unpredictable than training data, making it difficult for non-interpretable IFD systems to ensure reliable performance in practical operating conditions.
Therefore, research on interpretability holds significant scientific and industrial value for advancing IFD~\cite{norExplainableAIXAI2021}.

Interpretability refers to \textit{the ability to provide explanations in understandable terms to a human}~\cite{zhangSurveyNeuralNetwork2021}. It has flourished in computer vision (CV) and natural language processing (NLP) but is still in its early-stage in IFD~\cite{borghesaniFourierbasedexplanation1DCNNs2023}. Interpretability methods can be categorized along various dimensions. Based on usage scenarios, they can be divided into ante-hoc (active) and post-hoc (passive).

Ante-hoc interpretability requires targeted modifications before training, such as introducing additional structures or altering the training process~\cite{abidRobustInterpretableDeep2020,yuanLWNetinterpretablenetwork2022,liuNTScatNetinterpretableconvolutional2022}. For example, Li~\textit{et~al}.~\cite{liWaveletKernelNetInterpretableDeep2022} used wavelet transforms to design interpretable wavelet convolution kernels for extracting impulsive components. Zhang~\textit{et~al}.~\cite{zhangInterpretableLatentDenoising2024}. further extended this approach to generative models, enhancing data quality and interpretability. Wang~\textit{et~al}.~\cite{wangFullyinterpretableneural2022} utilized traditional feature extraction methods to design ELMs for localizing resonant frequency bands. Li~\textit{et~al}.~\cite{liUnderstandingimprovingdeep2019} optimized transformers by leveraging the multi-head attention mechanism to interpret contributions from different segments, with subsequent improvements made by others~\cite{yangInterpretingnetworkknowledge2020,tangSignalTransformerRobustInterpretable2022,liVariationalAttentionBasedInterpretable2024}. An~\textit{et~al}.~\cite{anInterpretableNeuralNetwork2022,anAdversarialAlgorithmUnrolling2024,zenginterpretablealgorithmunrolling2024} developed interpretable network structures through algorithmic unrolling, using learned dictionaries to highlight fault features.

While ante-hoc methods achieve unique interpretability, they also impose constraints on model architecture, limiting its flexibility and scalability. Moreover, despite claims of high diagnostic accuracy~\cite{chenTFNinterpretableneural2024,chenInterpretingwhattypical2024}, such approaches often lack optimization potential in complex scenarios and may even compromise diagnostic performance.

Conversely, post-hoc interpretability focuses on explaining pre-trained models without imposing additional constraints. This approach preserves the baseline model's performance while maintaining its flexibility and scalability. Current research in post-hoc interpretability primarily focuses on attribution, which involves \textit{assign credit (or blame) to the input features}~\cite{zhangSurveyNeuralNetwork2021}. Common attribution methods include gradient class activation mapping (grad-CAM)~\cite{selvarajuGradCAMVisualExplanations2017}, layer-wise relevance propagation (LRP),integrated gradients (IG)~\cite{sundararajanAxiomaticattributiondeep2017a}, and \underline{SH}apley \underline{A}dditive ex\underline{P}lanations (SHAP)~\cite{lundbergunifiedapproachinterpreting2017}, etc. Some researchers, inspired by advances in CV, have extended existing attribution methods to the IFD domain. For instance, Wu~\textit{et~al}.~\cite{wuhybridclassificationautoencoder2021} and Grezmak~\textit{et~al}.~\cite{grezmakInterpretableConvolutionalNeural2020} applied time-frequency transforms to convert vibration signals into 2D images and calculated contributions using CAM and LRP.
However, this approach achieved suboptimal interpretability and is not end-to-end.
To improve interpretability, Li~\textit{et~al}.~\cite{liWhiteningNetGeneralizedNetwork2021} and Li~\textit{et~al}.~\cite{liMultilayerGradCAMeffective2023} employed IG and an optimized Grad-CAM, respectively, to generate time-domain explanations, and use \textit{Fourier} transform for further analysis.

Although post-hoc methods excel in performance, flexibility, and scalability, their primary challenge in IFD lies in the explanation form. The outputs are typically in the time domain, aligning with the input of end-to-end IFD models. However, time-domain explanations often fail to directly reveal the fault information, as they primarily highlight moments of impact~\cite{chenExplainableDeepEnsemble2023}. Consequently, the explanations of existing post-hoc methods remains unsatisfactory, even when post-processing techniques are employed to indirectly convert these explanations into other domains~\cite{liWhiteningNetGeneralizedNetwork2021,liMultilayerGradCAMeffective2023}.

Recently, researchers have integrated domain transform with SHAP to extend explanations from the time domain to more informative domains without modifying the end-to-end architecture. These domains include the frequency domain~\cite{herwigExplainingdeepneural2023}, time-frequency domain~\cite{herwigExplainingdeepneural2023}, envelope domain~\cite{deckerDoesYourModel2023}, and cyclic-spectral domain~\cite{chenCSSHAPExtendingSHAP2025 }, offering richer insights into fault characteristics.
Specifically, they preprocess the data using domain transform and integrate the end-to-end model with the inverse transform, thereby leveraging the existing SHAP framework to obtain explanations in other domains. While this combination is conceptually straightforward, it has demonstrated remarkable interpretability in various experiments. Besides, the choice of SHAP among numerous attribution methods can be justified: 1) SHAP provides superior explanatory performance and is widely regarded as a benchmark for attribution~\cite{warneckeEvaluatingexplanationmethods2020}. 2) Unlike gradient-based methods (e.g, CAM and IG), SHAP is not influenced by gradient disruptions caused by domain transforms.

However, the computation of SHAP involves subset enumeration and data distribution expectation, making it extremely time-consuming~\cite{warneckeEvaluatingexplanationmethods2020}. The increased dimensionality introduced by domain transform further exacerbates this issue. Thus, enhancing SHAP's computational efficiency emerges as a critical challenge for post-hoc interpretability in IFD, alongside improving the form of explanations.

To address this challenge, we propose patch-wise attribution and SHEP. Patch-wise attribution groups adjacent features into a single patch, reducing feature dimensions at the cost of coarser explanation granularity. SHEP, on the other hand, simplifies the complex subset enumeration by selecting two representative cases to approximate the computationally expensive SHAP, reducing its complexity from exponential to linear. Together, these methods significantly improve post-hoc attribution efficiency, making real-time explanations feasible. In experiments, we assess the rationality of SHEP and demonstrate the advantages of the proposed methods in terms of both interpretability and efficiency. Besides, using cosine similarity as a metric, we further validate the reliability of approximating SHAP with SHEP.

To foster the development of the academic community, we will release the full code to ensure reproducibility. With superior interpretability, efficiency, and open-source code, the proposed SHEP has the potential to become a benchmark for post-hoc interpretability in the IFD domain. Our contributions can be summarized as follows:
\begin{enumerate}
        \item Patch-wise attribution is proposed  as an optional approach to improve attribution computational efficiency, at the cost of coarser explanation granularity.
        \item SHEP is introduced to approximate SHAP, reducing the computational complexity from exponential to linear.
        \item Extensive experiments validate the advantages of SHEP in both efficiency and interpretability, confirming its reliability in approximating SHAP.
        \item The open-source code ensures the authenticity and reproducibility of SHEP, advancing the academic community and positioning it as a potential universal tool for post-hoc interpretability in the IFD domain.
\end{enumerate}

\section{Preliminary}
\label{sec:Preliminary}

\subsection{SHAP}
SHAP~\cite{lundbergunifiedapproachinterpreting2017} is a classic post-hoc explanation method that quantifies the contribution of each feature in a sample to machine learning predictions, a process known as attribution. SHAP is based on the Shapley value from game theory. The Shapley value~\cite{shapleyvaluenpersongames1953} is a mathematical approach for determining fair and effective resource allocation within a group, widely used in profit distribution, cost sharing, and credit assignment.

Denoting the players as $p$, the union of players as $S$, and the value function as $v:S\rightarrow \mathbb{R}$. The marginal contribution $\Delta v (p,S)$, which measures the incremental value contributed by player $p$ joining union $S$, can be expressed as:
\begin{equation}
        \Delta v (p,S) = v ( S \cup\{p \} )-v (S).
\end{equation}
Then, the Shapley value of member $p_i$ is defined as the expected marginal contribution $\Delta v (p_i,S)$ to all possible subsets $S$ of the entire union $U=\{p_1,p_2,\cdots,p_d\}$:
\begin{equation}
        \psi(v,U)_{i}=\sum_{S \subseteq U \setminus\{p_i \}} \frac{s! ( d-s-1 )!} {d!} \cdot \Delta v  (p_i,S)
        \label{eq:SHAC_Shapley}
\end{equation}
where, $s$ represents the number of elements of $S$, and the weight ${s! ( n-s-1 )!} \,/\,{n!}$ represents the probability of player $p_i$ joining the subset $S$.

Shapley values can calculate the benefits of coalition members but are not directly applicable to machine learning. In Shapley value, the input dimension (i.e., the number of set members) of the value function $v$ can be different, whereas the input dimension for machine learning is fixed.

Thus, SHAP considers all feature dimensions as the entire union $U$, where $|U|=d$, and $d$ is the number of features. For a given subset $S$, features within $S$ retain their fixed values from the input sample $\tilde x$, while features outside $S$ are sampled from the data distribution $X$. The resulting samples can be expressed as:
\begin{equation}
        \tilde {\boldsymbol{x}}^S_{i} =\left\{ 
    \begin{array}{ll}
        \tilde {\boldsymbol{x}}_i \, \textrm{(constant)}, &\textrm{if}\; i \, \in\, S\\
        X_i \, \textrm{(variable)}, \quad &\textrm{if}\; i \, \notin\, S.
    \end{array}
\right.
\label{eq:SHAC_sample}
\end{equation}
Denoting the model as $\mathcal{M}$, the value function in SHAP can be expressed as:
\begin{equation}
        \begin{split}
        v_{\mathcal{M}, X, \tilde {\boldsymbol{x}}} ( S )&= \mathbb{E} [\mathcal{M} ( \tilde {\boldsymbol{x}}^S)]-\mathbb{E} [\mathcal{M} ( X )],\\
         &= \int\! \mathcal{M}(\tilde {\boldsymbol{x}}^S) \,\mathrm{d} \mathbb{P}_{X} - \int\! \mathcal{M}(X) \,\mathrm{d} \mathbb{P}_{X}.
        \end{split}
\end{equation}
It can be understood as the output difference of the model caused by constraining the dimensions within $S$ from the data distribution $X$ to the fixed values of sample $\tilde {\boldsymbol{x}}$. Substituting the value function $v$ into \eqref{eq:SHAC_Shapley}, the SHAP result $\psi_{\mathcal{M},X}(\tilde {\boldsymbol{x}}) _i$ is computed as:
\begin{equation}
        \begin{split}
       &\psi_{\mathcal{M},X}(\tilde {\boldsymbol{x}}) _i \!= \\
       &\sum_{S \subseteq U \setminus\{i \}} \! \frac{s! ( d-s-1 )!} {d!} \left( \mathbb{E} [\mathcal{M} ( \tilde {\boldsymbol{x}}^{S \cup \{i\}})]-\mathbb{E} [\mathcal{M} ( \tilde {\boldsymbol{x}}^S)]\right).
        \end{split}
        \label{eq:SHAC_SHAP}
\end{equation}
The authors of SHAP have provided a Python open-source package~\cite{lundbergshap2024} to facilitate SHAP's application.

SHAP computation consists of two components: estimating the expectation over the data distribution and enumerating subsets from the entire union $U$. 
The computational cost of distribution expectation is relatively small since it scales linearly with the number of background samples $n$, i.e., $\mathcal O(n)$. Additionally, it is often mitigated by selecting a smaller representative dataset for calculation. Conversely, subset enumeration is computationally expensive, growing exponentially with the dimensionality $d$ of the sample, i.e., $\mathcal O(2^d)$. Although permutation techniques can reduce the complexity to $\mathcal O(2k_pd)$, where $k_p$ is the number of permutations, the computational burden remains significant~\cite{lundbergunifiedapproachinterpreting2017,anconaExplainingdeepneural2019}.

Vibration samples used in IFD often exceed several thousand dimensions, making SHAP attribution computationally intensive. Thus, reducing SHAP's computational cost is both a critical and practical challenge in IFD applications.

\subsection{Combination of SHAP and Domain Transform in IFD}
\label{subsec:domain_transform}

End-to-end IFD models typically use time-domain signals as input, which often fail to directly capture fault-related information. Transforming signals from the time domain to a more informative domain is a common practice in signal processing-based fault diagnosis~\cite{randallVibrationBasedConditionMonitoring2021}. Inspired by this, some researchers have extended SHAP to other domains via domain transforms to provide clearer attribution explanations~\cite{deckerDoesYourModel2023,herwigExplainingdeepneural2023,chenCSSHAPExtendingSHAP2025}.

\begin{figure}[htbp]
        \centering
        \includegraphics[width=8.5 cm]{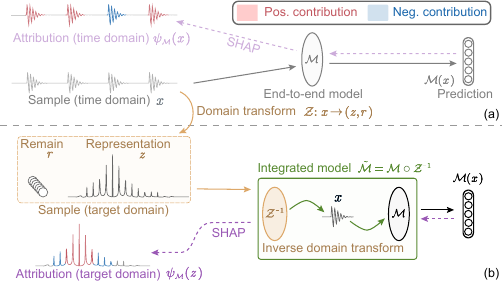}
        \caption{The illustration of SHAP attribution in the time and target domains. (a)~Traditional time-domain SHAP produces attribution results as components with varying impulse timings, offering limited clarity. (b)~By applying the domain transform, the input is preprocessed into the target domain, and the model integrates the inverse transform, enabling clear and interpretable attributions within the target domain.}
        \label{fig:2-DomainTransform}
\end{figure}

As shown in Fig.~\ref{fig:2-DomainTransform}(a), traditional time-domain SHAP computes the contribution of each segment of the input time-domain sample ${\boldsymbol{x}}$ to the prediction $\mathcal{M}({\boldsymbol{x}})$ using~\eqref{eq:SHAC_SHAP}, producing results at specific time points $\psi _{\mathcal{M}}({\boldsymbol{x}})$. While effective for signals with prominent fault features and low noise levels, this method faces challenges in real-world scenarios where high-noise signals often obscure fault features.

With the domain transform $\mathcal{Z}\!: {\boldsymbol{x}} \!\rightarrow\! ({\boldsymbol{z}},{\boldsymbol{r}})$ and inverse domain transform $\mathcal{Z}^{-1}\!: ({\boldsymbol{z}},{\boldsymbol{r}}) \!\rightarrow\! {\boldsymbol{x}}$, the process of extending SHAP to the target domain is depicted in Fig.~\ref{fig:2-DomainTransform}(b). It involves two key components: {\bfseries \itshape 1) Sample Preparation:} The domain transform $\mathcal{Z}$ converts the sample ${\boldsymbol{x}}$ from the time domain to the target domain (i.e., representation ${\boldsymbol{z}}$ for analysis and remain ${\boldsymbol{r}}$ for reconstruction). {\bfseries \itshape 2) Model Integration:}  The inverse domain transform $\mathcal{Z}^{-1}$ is integrated with the end-to-end model $\mathcal{M}$, allowing the target-domain sample to serve as the model input without altering the end-to-end structure. Finally, SHAP analysis is performed using the target-domain sample $({\boldsymbol{z}},{\boldsymbol{r}})$ and the integrated model $\tilde{\mathcal{M}}$, yielding the contributions of each component in the target domain $\psi _{\mathcal{M}}({\boldsymbol{z}})$.

Currently, the domain transforms applied to SHAP primarily include frequency (Freq) domain, envelope (Env) domain, time-frequency (TF) domain, and cyclic-spectral (CS) domain, as summarized in Table~\ref{tab:2-DomainTransform}. Each transform reveals distinct signal characteristics: the Freq domain emphasizes general frequencies, the Env domain highlights modulation frequencies, the TF domain captures signal characteristics across both frequency and time dimensions, and the CS domain identifies both carrier and modulation frequencies. These transforms provide unique insights, enabling fault diagnosis explanations tailored to specific signal properties.

\begin{table}[htbp]
        \centering
        \caption{The Calculation of Representations and Remains across Four Domain Transforms.}
        \label{tab:2-DomainTransform}
        \begin{threeparttable}
                \footnotesize
                \begin{tabular*}{\hsize}{@{\extracolsep{\fill}}ccc}
                        \toprule[1pt]
                        Domain                   & Representation\tnote{a} ($z$)         & Remain\tnote{b} ($r$)        \\ \toprule[1pt]
                        Freq &   \makecell[c]{$X(f)\!=\!\int x(t) \mathrm{e}^{-\mathrm{j}2\pi ft} \,{\mathrm d}t$, \\ $z\!=\!|X(f)|^2$} & \makecell[c]{$r\!=\!\mathrm{ang}[X(f)]$} \\ \midrule[0.3pt]
                        Env & \makecell[c]{$z(t)\!=\!x(t)+\mathrm{j}\hat x(t),\; \hat x(t) \!=\! \frac{1}{\pi} \int \frac{x(\tau)}{t-\tau} \,{\mathrm d}t$,\\ $Z(f)\!=\!\int \left( |z(t)| - \overline{|z(t)|} \right) \mathrm{e}^{-\mathrm{j}2\pi ft} \,{\mathrm d}t$, \\ $z\!=\!|Z(f)|^2$} & \makecell[c]{$r_1\!=\!\mathrm{ang}(z(t))$,\\ $r_2\!=\!\overline{|z(t)|}$,\\ $r_3\!=\!\mathrm{ang}[Z(f)]$} \\ \midrule[0.3pt]
                        TF & \makecell[c]{$Z(f,t)\!=\!\int x(\tau-t)w(\tau) \mathrm{e}^{-\mathrm{j}2\pi f \tau} {\mathrm d}\tau$,\\ $z \!=\! |Z(f,t)|^2$} & \makecell[c]{$r\!=\!\mathrm{ang}[Z(f,t)]$} \\ \midrule[0.3pt] 
                        CS & \makecell[c]{$Z(f,t)\!=\!\int x(\tau-t)w(\tau) \mathrm{e}^{-\mathrm{j}2\pi f \tau} {\mathrm d}\tau$,\\ $Z(f,\alpha)\!=\!\int |Z(f,t)|^2 \mathrm{e}^{-\mathrm{j}2\pi \alpha t} {\mathrm d} t$,\\ $z \!=\! |Z(f,\alpha)|^2$} & \makecell[c]{$r_1\!=\!\mathrm{ang}[Z(f,t)]$,\\$r_2\!=\!\mathrm{ang}[Z(f,\alpha)]$} \\ \bottomrule[1pt]
                \end{tabular*}
                \begin{tablenotes}
                        \footnotesize
                        \item[a] $w(t)$ represents the window function used in Short-Time \textit{Fourier} Transform.
                        \item[b] $\mathrm{ang}(\cdot)$ denotes the phase angle.
                \end{tablenotes}
        \end{threeparttable}
\end{table}

However, achieving higher resolution in 2D domains (i.e., TF and CS) notably increases data dimensionality compared to the original time domain. This rise in dimensionality $d$ substantially amplifies SHAP's computational cost for subset enumeration, underscoring the critical need to reduce computational cost in SHAP combined with domain transforms.

\section{Method}

As analyzed in Section~\ref{sec:Preliminary}, the computational cost of SHAP primarily lies in subset enumeration, and the increased feature dimensionality caused by 2D domain transforms exacerbates this issue. To address this, we propose patch-wise attribution and SHapley Estimated Explanation (SHEP). Specifically, patch-wise attribution reduces feature dimensionality $d$ by grouping adjacent sample points (1D) or pixels (2D) into cohesive patches, while SHEP approximates SHAP by evaluating only two representative cases (i.e., {\itshape remove} and {\itshape add}), reducing the computational complexity from $\mathcal{O}(2^d)$ to $\mathcal{O}(d)$.

\subsection{Patch-Wise Attribution}

As shown in~\eqref{eq:SHAC_SHAP}, the complexity of subset enumeration in SHAP is $\mathcal{O}(2^{d})$. A natural solution to reduce computational costs is to lower the feature dimensionality $d$ of the sample. To achieve this, we group multiple adjacent points or pixels into a single cohesive patch and calculate their joint contribution, i.e., patch-wise attribution.

\begin{figure}[htbp]
        \centering
        \includegraphics[width=8.5 cm]{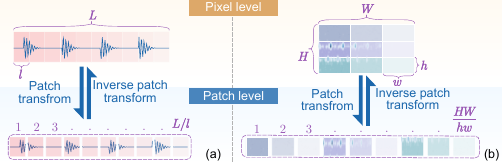}
        \caption{The process of patch transform for 1D and 2D samples. (a)~1D sample (i.e., Freq and Env domains). (b)~2D sample (i.e., TF and CS domains).}
        \label{fig:3-PatchwiseAtrribution}
\end{figure}

As depicted in Fig.~\ref{fig:3-PatchwiseAtrribution}, each pixel in the original sample represents an individual dimension. The patch transform $\mathcal{P}$ aggregates adjacent pixels into patches of a defined size, transforming the sample into a patch-wise representation ${\boldsymbol{p}}$:
\begin{equation}
        \mathcal{P}({\boldsymbol{x}}) = {\boldsymbol{p}}.
\end{equation}
where each patch serves as a single dimension encompassing multiple pixels. Furthermore, the patch transform is fully reversible. The inverse patch transform $\mathcal{P}^{-1}$ can seamlessly reconstruct the original sample by concatenating the patches:
\begin{equation}
        \mathcal{P}^{-1}({\boldsymbol{p}}) = {\boldsymbol{x}}.
\end{equation}
The usage of patch-wise attribution is consistent with the domain transform $\mathcal{Z}$ described in Section~\ref{subsec:domain_transform}. Specifically, the patch transform $\mathcal{P}$ is employed for sample preparation, while the inverse patch transform $\mathcal{P}^{-1}$ is integrated with the model.

Thus, patch-wise attribution reduces the dimension $d$ during the attribution process, thereby accelerating computation. Admittedly, this speed enhancement comes at the expense of coarser granularity in attribution results. Denoting the patch size as $k$, the computational complexity of patch-wise attribution becomes $\mathcal{O}({2^{d/k}})$, while the granularity of the attribution results is $k$. On one hand, the exponential increase in computational efficiency outweighs the linear trade-off in granularity, making patch-wise attribution a highly competitive solution. On the other hand, adjusting the patch size is crucial to balancing computational cost and explanation granularity, ensuring the approach aligns with practical requirements.

\subsection{SHEP}
\label{subsec:SHEP}
SHAP requires the complete subset enumeration of all dimensions, which prompts us to question the necessity of full enumeration. Specifically, whether satisfactory attribution results can be  achieved by considering only a few typical cases. To address this, SHEP is proposed as an approximation to SHAP. It consists of two key cases: SHEP-\textit{Remove} and SHEP-\textit{Add}.

\begin{figure}[htbp]
        \centering
        \includegraphics[width=8.5 cm]{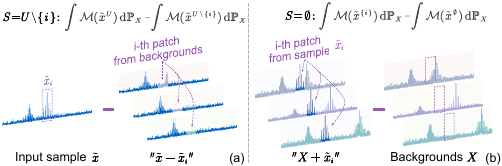}
        \caption{The illustration of SHEP-\textit{Remove} and SHEP-\textit{Add}. (a)~SHEP-\textit{Remove} calculates the contribution by "\textit{removing}" the feature $\tilde {\boldsymbol{x}}_i$ from the input sample $\tilde {\boldsymbol{x}}$. (b)~SHEP-\textit{Add} calculates the contribution by "\textit{adding}" the feature $\tilde {\boldsymbol{x}}_i$ to the data distribution $X$.}
        \label{fig:3-SHEP}
\end{figure}

As shown in Fig.~\ref{fig:3-SHEP}(a), the concept of SHEP-\textit{Remove} aligns with mainstream perturbation-based attribution methods, where a specified feature (or patch) $\tilde {\boldsymbol{x}}_i$ is removed from the current sample $\tilde {\boldsymbol{x}}$, and the feature's contribution is measured based on the resulting change in the model's output. Unlike common techniques such as masking, blurring, or scaling, the feature removal operation follows the SHAP framework depicted in \eqref{eq:SHAC_sample}. Specifically, removing feature $\tilde {\boldsymbol{x}}_i$ means degrading it to the data distribution $X_i$ and obtaining the analysis samples $\tilde {\boldsymbol{x}}^{U\setminus \{i\}}$, where $U$ represents the complete set of feature dimensions. Thus, the computation of the SHEP-\textit{Remove} attribution $\psi_{\mathcal{M},X}^{\mathrm{Rm}}(\tilde {\boldsymbol{x}})$ can be expressed as:
\begin{equation}
        \begin{split}
                \psi_{\mathcal{M},X}^{\mathrm{Rm}}(\tilde {\boldsymbol{x}})_i &= \mathcal{M} ( \tilde {\boldsymbol{x}}) - \mathbb{E} \left[\mathcal{M} \left( \tilde {\boldsymbol{x}}^{U\setminus \{i\}} \right)\right]\\
                &= \mathbb{E} \left[\mathcal{M} ( \tilde {\boldsymbol{x}}^{U})\right]-\mathbb{E} \left[\mathcal{M} \left( \tilde {\boldsymbol{x}}^{U\setminus \{i\}}\right)\right].
         \end{split}
         \label{eq:SHEC_Remove}
\end{equation}
It is evident that SHEP-\textit{Remove} is a special case of the SHAP computation in \eqref{eq:SHAC_SHAP}, where $S=U\setminus \{i\}$.

Unlike SHEP-\textit{Remove}, which is based on the current sample's perspective, SHEP-\textit{Add} adopts the perspective of the data distribution. As shown in Fig.~\ref{fig:3-SHEP}(b), SHEP-\textit{Add} adds feature $\tilde {\boldsymbol{x}}_i$ to the background sample $X$, which is drawn from the data distribution, thus obtaining the analysis samples $\tilde {\boldsymbol{x}}^{\{i\}}$. The contribution of this feature $\psi_{\mathcal{M},X}^{\mathrm{Add}}(\tilde {\boldsymbol{x}})_i$ is then measured by the change in the model's output:
\begin{equation}
        \begin{split}
                \psi_{\mathcal{M},X}^{\mathrm{Add}}(\tilde {\boldsymbol{x}})_i &= \mathbb{E} \left[\mathcal{M} \left( \tilde {\boldsymbol{x}}^{\{i\}} \right) - \mathcal{M}(X)\right]\\
                &= \mathbb{E} \left[\mathcal{M} ( \tilde {\boldsymbol{x}}^{\{i\}})\right]-\mathbb{E} \left[\mathcal{M} \left( \tilde {\boldsymbol{x}}^{\emptyset}\right) \right].
         \end{split}
         \label{eq:SHEC_Add}
\end{equation}
SHEP-\textit{Add} is also a special case of \eqref{eq:SHAC_SHAP} with $S=\emptyset$.

Finally, the output of SHEP can be expressed as the combination of these two cases:
\begin{equation}
        \psi_{\mathcal{M},X}^{\mathrm{SHEP}}(\tilde {\boldsymbol{x}})_i = \frac{1}{2} \left( \psi_{\mathcal{M},X}^{\mathrm{Rm}}(\tilde {\boldsymbol{x}})_i + \psi_{\mathcal{M},X}^{\mathrm{Add}}(\tilde {\boldsymbol{x}})_i \right).
        \label{eq:SHEP}
\end{equation}

\begin{figure*}[t]
        \centering 
        \includegraphics[width=17.5 cm]{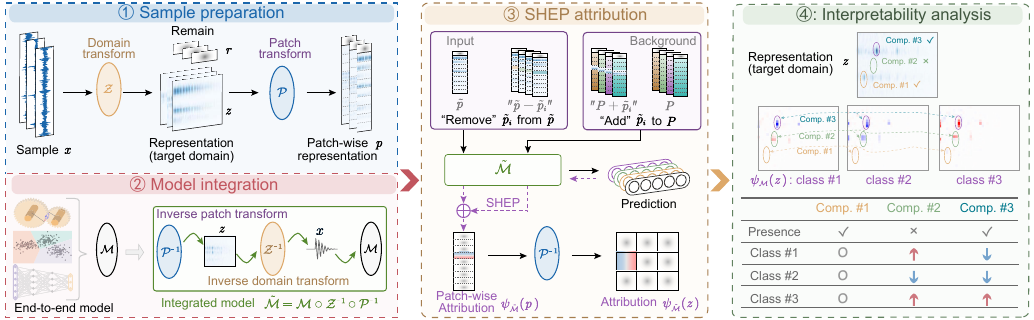}
        \caption{The flow chart of applying SHEP for interpretability analysis of IFD model.}
        \label{fig:3-Framework}
\end{figure*}

However, given the vast possible subsets in SHAP, why do we choose these two cases (SHEP-\textit{Remove} and SHEP-\textit{Add})? First, we need to clarify the logic behind attribution. A sample's positive contribution to the predicted class ${\boldsymbol{y}}_j$ arises not only from the presence of features specific to class ${\boldsymbol{y}}_j$, but also from the absence of features associated with other classes. An effective attribution method must account for both factors simultaneously. In fact, due to the performance advantages of neural networks, their predictions are typically sparse. Consequently, the analysis samples formed by \eqref{eq:SHAC_sample} often breaks this sparsity, leading to a significant decrease in the dominant class and an indistinguishable increase in other classes, rather than a simple combination of the corresponding prediction tensors.

Therefore, SHEP-\textit{Remove}, based on the current sample's perspective, is better at capturing the influence of the presence of features within the sample, aligning with traditional perturbation-based attribution methods. On the other hand, SHEP-\textit{Add}, which operates from the background sample's perspective, is more effective at capturing the impact of the absence of features present in the background sample. Thus, SHEP combines the strengths of both approaches, providing a more comprehensive assessment of feature contributions. Additionally, we will further analyze these methods to validate this hypothesis in Section~\ref{subsec:SHEPAnalysis}.

\subsection{Framework}

Domain transform, patch transform, and SHEP can be combined as an effective and efficient solution for attribution explanations of end-to-end models in IFD. Specifically, domain transform extends attribution explanations from the time domain to more informative domains, enhancing clarity. Meanwhile, patch transform reduces feature dimensions, and SHEP simplifies computational complexity, thereby accelerating the calculation process.

The complete process of applying SHEP for interpretability analysis of end-to-end models in IFD is shown in Fig.~\ref{fig:3-Framework}, and consists of four parts: sample preparation, model integration, SHEP attribution, and interpretability analysis. First, the input samples ${\boldsymbol{x}}$ are preprocessed using domain transform $\mathcal{Z}$ and patch transform $\mathcal{P}$ sequentially to obtain patch-wise representations ${\boldsymbol{p}}$. Meanwhile, inverse domain transform $\mathcal{Z}^{-1}$, inverse patch transform $\mathcal{Z}^{-1}$, and the end-to-end model $\mathcal{M}$ are combined to form the integrated model $\tilde{\mathcal{M}}$. In SHEP attribution, patch-wise attributions $\psi_{\tilde{\mathcal{M}}}({\boldsymbol{p}})$ are computed using \eqref{eq:SHEC_Remove}-\eqref{eq:SHEP} and target domain attributions $\psi_{\tilde{\mathcal{M}}}({\boldsymbol{z}})$ are obtained through inverse patch transform $\mathcal{Z}^{-1}$. Finally, interpretability analysis is performed based on the attribution results, where red represents positive contributions and blue represents negative contributions.

Notably, the attribution results are influenced by two factors: the relevance of signal components to the current class and their presence. Specifically, the presence (absence) of fault components related to the current class leads to a positive (negative) contribution, while the presence (absence) of fault components related to other classes results in a negative (positive) contribution. As shown in the interpretability analysis of Fig.~\ref{fig:3-Framework}: \textbf{Component \#1}, unrelated to all three classes, has zero contribution across all classes regardless of its presence. 
\textbf{Component \#2}, related to Class \#2, its absence contributes negatively to Class \#2 and positively to Classes \#1 and \#3. 
\textbf{Component \#3}, related to Class \#3, its presence contributes positively to Class \#3 but negatively to Classes \#1 and \#2.

\section{Simulation and Detailed Analysis}

The simulation dataset has fully known fault logic, facilitating the evaluation of interpretability that is challenging to achieve with other datasets due to the lack of ground-truth. Therefore, we conduct a series of detailed analyses on this simulation dataset. First, we discuss the effect of patch-wise attribution, followed by validation of the hypothesis of SHEP-\textit{Add} and SHEP-\textit{Remove}. Finally, we compare the explanation results and computational efficiency with other methods.

For the predictive model $\mathcal{M}$, we selected an end-to-end convolutional neural network (CNN) for analysis, with its architecture detailed in Table~\ref{tab:Exp-modelarch}.
\begin{table}[htbp]
        \centering
        \caption{The Model Architecture Used in Experiments.\label{tab:Exp-modelarch}}
        \begin{threeparttable}
                \footnotesize
                \begin{tabular*}{\hsize}{@{\extracolsep{\fill}}lll}
                        \toprule[1pt]
                        No. & Basic unit                                    & Output size     \\
                        \midrule[0.3pt]
                        -         & Input                                    & 1$\times$2000  \\
                        1         & Conv(8@7)\tnote{a}-BN-ReLU-MaxPool(2)       & 8$\times$997   \\
                        2$\sim$7:$\rightarrow i$  & {[}Conv($\textrm{2}^{i\textrm{+2}}$@3)-BN-ReLU-MaxPool(2){]}*6 & 512$\times$13  \\
                        9         & Conv(1024@3)-BN-ReLU-AdapMaxPool(1) & 1024$\times$1  \\
                        10        & Flatten-FC(256)-ReLU-FC(64)-ReLU-FC($K$) & $K$\tnote{b} \\
                        \bottomrule[1pt]
                \end{tabular*}

                \begin{tablenotes}
                        \smallskip
                        \footnotesize
                        \item[a] Conv(\textit{x}@\textit{y}): represents a convolutional layer with \textit{x} output channels and a kernel size of \textit{y}. Besides, the stride is 1, the padding is 0, and the input channels could be determined by the output size of the previous layer.
                        \item[b] $K$: represents the number of classes in the dataset.
                \end{tablenotes}
        \end{threeparttable}
\end{table}

\subsection{Simulation Dataset}

The simulated samples $x$ are constructed as the combination of several periodic-impulse components $c(t)$ and noise $n(t)$:
\begin{equation}
        x=\sum_{i=1}^{2} A \cdot c^i(t)+n(t)
\end{equation}
where, $A\sim\mathcal{U} (0.8,1)$ is the amplitude coefficient, and $\mathcal{U} (a,b)$ represents a uniform distribution with a lower bound $a$ and an upper bound $b$. $c^i(t)$ is the $i$-th periodic-impulse component, and $n(t)$ represents Gaussian white noise with a signal-to-noise ratio (SNR) of 0.

The periodic-impulse component $c(t)$ could be denoted as:
\begin{equation}
        \begin{split}
                c_{f_m,f_c}(t)\!=\!\sum_{k\in \mathbb{N}} \mathrm{e}^{-\beta(t-k/f_m)} \sin \big( 2\pi f_c (t-\frac{k}{f_m})+\phi \big)
        \end{split}
\end{equation}
where, $f_m$ is the excitation frequency of the fault impulse, $f_c$ is the response frequency, $\beta=0.04$ is the damping, and $\phi \sim \mathcal{U} (0,2\pi)$ is the initial phase.
Different periodic-impulse components $c(t)$ have different $f_m$ and $f_i$.

\begin{figure}[htbp]
        \centering
        \includegraphics[width=8.5 cm]{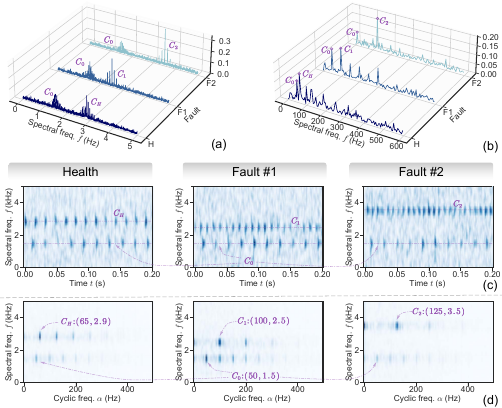}
        \caption{The domain representations of three classes in the simulation dataset. (a)~Frequency domain. (b)~Envelope domain. (c)~Time-frequency domain. (d)~Cyclic-spectral domain.}
        \label{fig:4-1-signal}
\end{figure}

In this dataset, the sampling frequency is set to 10 kHz, and three fault classes are defined: Health (H), Fault \#1 (F1), and Fault \#2 (F2). Their relationships to the components and corresponding parameters are detailed in Table~\ref{tab:ExpSimulation-data}. Each fault class contains two periodic-impulse components. Notably, Component $C_0$ is shared across all three classes and has zero contribution to any of them. In contrast, other components (i.e., $C_H$, $C_1$, $C_2$) are exclusive to a single class, contributing positively to their corresponding class while negatively impacting others. For better clarity, the multi-domain representations of these three fault classes are shown in Fig.~\ref{fig:4-1-signal}.

\begin{table}[htbp]
        \centering
        \caption{The Relationship between Fault Classes and Periodic-Impulse Components, along with Their Corresponding Parameters. \label{tab:ExpSimulation-data}}
        \begin{threeparttable}
                \footnotesize
                \begin{tabular*}{\hsize}{@{\extracolsep{\fill}}cccccc}
                \toprule[1pt]
                Component & Health & Fault \#1 & Fault \#2 & $f_c$ (kHz) & $f_m$ (Hz) \\
                \midrule[0.3pt]
                $C_0$     & \checkmark & \checkmark   & \multicolumn{1}{c|}{\checkmark} & 1.5 & 50    \\
                $C_H$   & \checkmark &          &   \multicolumn{1}{c|}{}   & $\mathcal{U}(1,4)$ & $\mathcal{U}(20,200)$ \\
                $C_1$  &        & \checkmark   &     \multicolumn{1}{c|}{}  & 2.5 & 100\\
                $C_2$  &        &          & \multicolumn{1}{c|}{\checkmark}  & 3.5 & 125\\
                \bottomrule[1pt]
                \end{tabular*}
        \end{threeparttable}
\end{table}

This experiment is set up as a 3-class classification task, with each class consisting of 5,000 samples, each of length 2,000, and normalized by mean-std normalization. Of these, 70\% are randomly selected for training, while the remaining 30\% are used for testing. The training parameters include 20 epochs, a batch size of 64, the Adam optimizer, and a learning rate of 0.001 with a 0.99 decay per epoch. After training, the CNN model achieved a test accuracy of 99.98\%.

\subsection{Analysis of Patch-Wise Attribution}
Compared to traditional attribution methods, patch-wise attribution significantly reduces computational complexity by lowering the feature dimensions, albeit at the cost of the explanation granularity. The relationship between patch size and efficiency will be discussed in Section~\ref{subsec:Efficiency}. Here, we only focus on the impact of patch size on granularity.

As shown in Table~\ref{tab:4-2-PatchSize}, we define the patch sizes at various levels for four different domains. In general, the higher the patch level, the larger the patch size and the fewer the feature dimensions.

\begin{table}[htbp]
        \centering
        \caption{The Patch Size Settings and the Corresponding Dimensions.}
        \label{tab:4-2-PatchSize}
        \begin{threeparttable}
                \footnotesize
                \begin{tabular*}{\hsize}{@{\extracolsep{\fill}}ccccc}
                        \toprule[1pt]
Patch                    & Freq   & Env    & TF         & CS        \\ \midrule[0.3pt]
\multicolumn{1}{c|}{-\tnote{a}} & 1\&1001\tnote{b}    & 4\&120     & 1\&26$\times$205     & 2\&26$\times$103    \\
\multicolumn{1}{c|}{\#1} & 335-(3)\tnote{c} & 124-(1) & 1067-(1,5)\tnote{d} & 912-(1,3) \\
\multicolumn{1}{c|}{\#2} & 168-(6) & 60-(2)  & 534-(2,5)  & 457-(2,3) \\
\multicolumn{1}{c|}{\#3} & 84-(12) & 34-(4)  & 274-(2,10) & 236-(2,6) \\
\multicolumn{1}{c|}{\#4} & 43-(24) & 19-(8)  & 148-(2,20) & 128-(4,6) \\
\multicolumn{1}{c|}{\#5} & 22-(48) & 12-(16) & 78-(4,20)  & 65-(4,12) \\
                        \bottomrule[1pt]
                \end{tabular*}
                \begin{tablenotes}
                        \footnotesize
                        \item[a] -: denotes the original input size without patch transform.
                        \item[b] \textit{p}\&\textit{q}: denotes the dimensions of remain $r$ and reprensetation $z$ are \textit{p} and \textit{q}, respectively.
                        \item[c] \textit{x}-(\textit{l}): denotes the output dimension is \textit{x}, under the patch size of \textit{l} in 1D domain.
                        \item[d] \textit{x}-(\textit{h},\textit{w}): denotes the height and width of patch is \textit{h} and \textit{w}, respectively, in 2D domain.
                \end{tablenotes}
        \end{threeparttable}
\end{table}

\begin{figure}[htbp]
        \centering
        \includegraphics[width=8.5 cm]{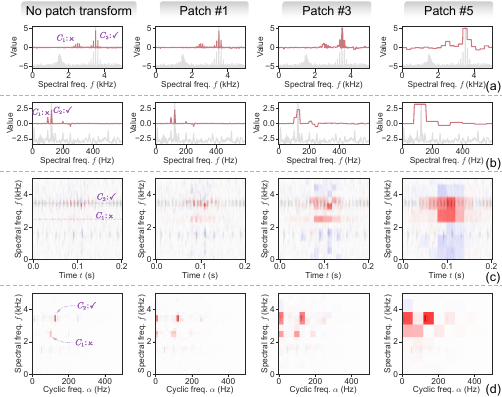}
        \caption{The SHEP attribution results in four domains for F2-prediction $y_2$ using a F2-class sample as the input $\tilde x$ in the simulation dataset with four patch sizes. (a)~Frequency domain. (b)~Envelope domain. (c)~Time-frequency domain. (d)~Cyclic-spectral domain.}
        \label{fig:4-3-SimuPatch}
\end{figure}

The SHEP results for different patch sizes across various domains are shown in Fig.~\ref{fig:4-3-SimuPatch}. Using F2-class samples as input, the contribution of each patch to the F2-prediction $y_2$ is measured. The results indicate that attributions in different domains emphasize distinct features, such as spectral frequency $f$, cyclic frequency $\alpha$, and time point $t$. The choice of domain thus depends on the specific requirements of the task, which is beyond the scope of our current work.

Regarding the effect of patch size, as the patch level increases, the explanation granularity in all domains becomes significantly coarser, though the overall attribution results remain correct. Specifically, the model accurately attributes the contribution to $y_2$ to the presence of $C_2$ ($C_2\!:\!\checkmark$) and the absence of $C_1$ ($C_1\!:\!\usym{2613}\,$), while the presence of $C_0$ contributes nothing. However, the coarser granularity makes it more challenging to distinguish between the contributions of different components. For example, with Patch \#5, the contributions of $C_2\!:\!\checkmark$ and $C_1\!:\!\usym{2613}\,$ become indistinguishable, while these components can still be effectively differentiated in Patch \#1 and even Patch \#3 scenarios.

In summary, as the patch size increases, the explanation granularity becomes coarser. While this does not affect the correctness of the attribution results, it makes distinguishing the contributions of different components more challenging. This suggests that excessively large patch sizes should be avoided, as they may degrade the quality of the explanation results.

\subsection{Analysis of SHEP-Remove and SHEP-Add}
\label{subsec:SHEPAnalysis}

\begin{figure}[htbp]
        \centering
        \includegraphics[width=8.5 cm]{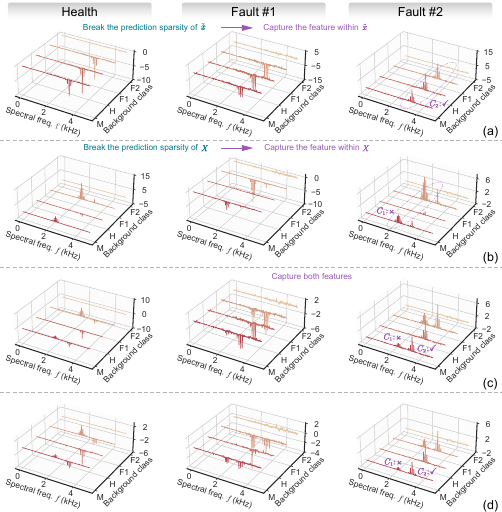}
        \caption{The attribution results of four attribution methods for different prediction classes using Fault \#2 as the input sample $\tilde {\boldsymbol{x}}$. Considering that background samples of different classes exhibit distinct data distributions, they are treated as the y-axis for seperate analysis, and ``\textit{M}'' represents the mean result across all background samples. (a)~SHEP-\textit{Remove}, (b)~SHEP-\textit{Add}, (c)~SHEP, and (d)~SHAP.}
        \label{fig:4-2-Combine}
\end{figure}

We have discussed the effect of SHEP-\textit{Remove} and SHEP-\textit{Add} in Section~\ref{subsec:SHEP}, and now we aim to validate it. Using a F2-class sample as input, the frequency domain attribution results of four methods are shown in Fig.~\ref{fig:4-2-Combine}. For better understanding, we analyze the classes of the background samples $X$ along the y-axis separately, and use ''\textit{M}'' to represent the mean of the results across all background samples, which corresponds to the actual attribution result.

As shown in Fig.~\ref{fig:4-2-Combine}(a), SHEP-\textit{Remove} attributes from the perspective of $\tilde{\boldsymbol{x}}$. The sample $\tilde{\boldsymbol{x}}$ contains the shared component $C_0$ and the unique component $C_2$. Additionally, its prediction for class $y_2$ exhibits sparsity. As for background class, background samples of the same class F2 do not contribute, while samples from different classes (H and F1) have contribution, which aligns with our expectations. As for frequency $f$, the feature exchange of $C_2$ breaks its sparse prediction for F2, significantly lowering ${\boldsymbol{y}}_2$ while increasing the probabilities for other classes indiscriminately. In contrast, feature exchanges of other frequencies (including 
$C_0$ and $C_1$) has almost no effect on the prediction result. Therefore, SHEP-\textit{Remove} excels in capturing the contribution of features within $\tilde{\boldsymbol{x}}$, e.g., $C_2$.

Conversely, SHEP-\textit{Add} attributes from the perspective of the background samples $X$, as shown in Fig.~\ref{fig:4-2-Combine}(b). The background samples from different classes exhibit distinct behaviors. For the H-class  background samples, the unique component $C_H$ is random, resulting in negligible contributions. For the F1-class background samples, the feature exchange of $C_1$ similarly breaks its sparse prediction for ${\boldsymbol{y}}_1$, thus contributing significantly. The F2-class background samples share the same class as the input sample $\tilde{\boldsymbol{x}}$, and have no contribution. Overall, SHEP-\textit{Add} excels in capturing the  influence of features absent in $x$ but present in $X$, e.g., $C_1$.

The results of SHEP are shown in Fig.~\ref{fig:4-2-Combine}(c). SHEP combines the advantages of both SHEP-\textit{Remove} and SHEP-\textit{Add}, effectively capturing the impact of both the presence of features within $\tilde{x}$ and the absence of features within $X$. Furthermore, the results of SHEP align well with the original SHAP, as shown in Fig.~\ref{fig:4-2-Combine}(d).
This indicates that SHEP not only demonstrates excellent attribution performance but also serves as an effective approximation of SHAP.

\subsection{Comparison of Interpretability}
Common attribution methods can be categorized into model-simplification-based methods (e.g., LIME~\cite{ribeiroWhyshouldtrust2016b}), gradient-based methods (e.g., Grad-CAM~\cite{selvarajuGradCAMVisualExplanations2017}, SmoothGrad~\cite{smilkovSmoothGradremovingnoise2017}), and perturbation-based approaches. However, model simplification methods are not suitable for heterogeneous patch-wise samples, and gradient-based methods are incompatible with domain transforms that may disrupt gradient propagation. To ensure fairness in comparison, we select perturbation-based attribution methods (i.e., Mask~\cite{zeilerVisualizingunderstandingconvolutional2014} and Scale~\cite{gwakRobustExplainableFault2023})
as the baseline for comparison, along with the original SHAP (using permutation mode with a set permutation number of $k_p=5$).

\begin{figure}[htbp]
        \centering
        \includegraphics[width=8.5 cm]{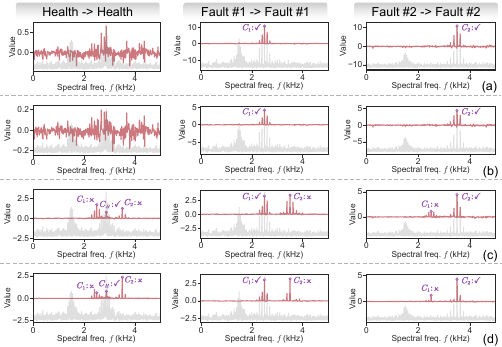}
        \caption{The frequency domain attribution results of different class samples for the corresponding prediction classes in the simulation dataset under four attribution methods with Patch \#1. Where, red color is the contribution, and gray color is input sample representation. (a)~Mask. (b)~Scale. (c)~SHEP. (d)~SHAP.}
        \label{fig:4-3-SimuAttrFreq}
\end{figure}

\begin{figure}[htbp]
        \centering
        \includegraphics[width=8.5 cm]{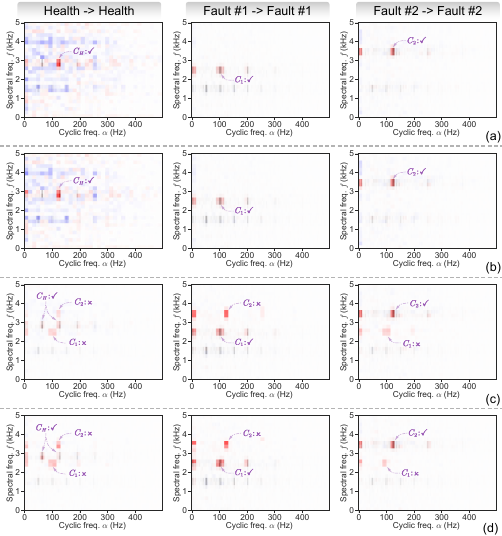}
        \caption{The cyclic-spectral domain attribution results of different class samples for the corresponding prediction classes in the simulation dataset under four attribution methods with Patch \#1. Among them, red color is positive contribution, blue color is negative contribution, and gray color is input sample representation. (a)~Mask. (b)~Scale. (c)~SHEP. (d)~SHAP.}
        \label{fig:4-3-SimuAttrCS}
\end{figure}

The attribution results of different methods in the frequency and cyclic-spectral domains are shown in Fig.~\ref{fig:4-3-SimuAttrFreq} and Fig.~\ref{fig:4-3-SimuAttrCS}, respectively. 
As shown in Fig.~\ref{fig:4-3-SimuAttrFreq}-\ref{fig:4-3-SimuAttrCS}~(a)-(b), both Mask and Scale, which are perturbation-based methods, exhibit similar behavior to the SHAP-Remove discussed in Section~\ref{subsec:SHEPAnalysis}.  These methods focus solely on the contribution of the components present in the current class, while neglecting the contribution arising from the absence of components from other classes.

In contrast, both SHAP and SHEP, as seen in Fig.~\ref{fig:4-3-SimuAttrFreq}-\ref{fig:4-3-SimuAttrCS}~(c)-(d), effectively capture both types of contributions, yielding more comprehensive attribution results. Notably, due to the randomness within component $C_0$ associated with the H-class, the Mask and Scale methods struggle to capture its contribution, while SHEP and SHAP are still able to weakly reflect it, demonstrating stronger attribution capabilities.

Although the visualizations in Fig.~\ref{fig:4-3-SimuAttrFreq}-\ref{fig:4-3-SimuAttrCS} are very clear, they do not comprehensively evaluate the interpretability performance of different methods. Hence, we use the cosine similarity between the results of each method and the SHAP result for quantitative evaluation. The calculation of cosine similarity $d_{\mathrm{cos}}$ can be expressed as:
\begin{equation}
        d_{\mathrm{cos}}(\boldsymbol{p},\boldsymbol{q})=\frac{\boldsymbol{p}\cdot \boldsymbol{q}}{|\boldsymbol{p}|\times|\boldsymbol{q}|}=\frac{\sum_i \boldsymbol{p}_i \boldsymbol{q}_i}{\sqrt{\sum_i \boldsymbol{p}_i^2} \sqrt{\sum_j \boldsymbol{q}_j^2}}.
\end{equation}
It is also applicable for calculating 2D images through a flattening operation.

\begin{figure}[htbp]
        \centering
        \includegraphics[width=8.5 cm]{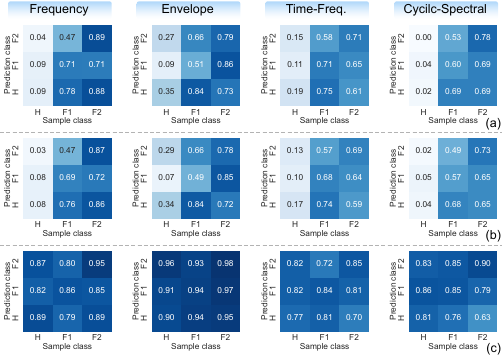}
        \caption{The cosine similarity matrices of different methods across various domains in the simulation dataset, with Patch \#1. Where, rows and columns represent different sample classes and prediction classes, respectively. (a)~Mask. (b)~Scale. (c)~SHEP.}
        \label{fig:4-3-SimuSimiMatrix}
\end{figure}

The cosine similarities for different domains and methods under Patch \#1 are shown in Fig.~\ref{fig:4-3-SimuSimiMatrix}. Among them, Mask and Scale exhibit similar performance, with their cosine similarities significantly lower than that of SHEP. This is because these methods can only capture the impact of feature presence within the current sample, whereas SHEP can also account for the absence of features from other classes, resulting in a closer correspondence with SHAP. Regarding sample classes, Mask and Scale perform poorly on H-class samples due to the randomness of $C_0$, as discussed in Fig.~\ref{fig:4-3-SimuAttrFreq}-\ref{fig:4-3-SimuAttrCS}~(a)-(b). Regarding domains, all methods show lower cosine similarities in the 2D domains (i.e., TF and CS) compared to the 1D domains (i.e., Freq and Env). This is because the 2D domains have higher dimensionality, which inherently results in lower similarity values.

\begin{figure}[htbp]
        \centering
        \includegraphics[width=8.5 cm]{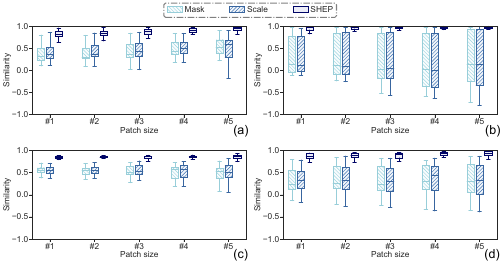}
        \caption{The statistic result of cosine similarity under different attribution methods and domains in the simulation dataset. (a)~Frequency domain. (b)~Envelope domain. (c)~Time-frequency domain. (d)~Cyclic-spectral domain.}
        \label{fig:4-3-SimuSimuStatistic}
\end{figure}

Fig.~\ref{fig:4-3-SimuSimuStatistic} presents a more comprehensive statistical result by compressing the class matrix and introducing the patch size axis based on Fig.~\ref{fig:4-3-SimuSimiMatrix}. As the patch size increases, the cosine similarity of SHEP significantly rises, indicating that SHEP's explanations align more closely with SHAP at lower granularities. In contrast, the mean similarities of Mask and Scale methods show little change, with even an increase in variance. Regarding domains, the similarity in the 1D domain is notably higher than in the 2D domain. Furthermore, the similarity variance in the Env and CS domains is significantly greater than in the Freq and TF domains, which may be attributed to the complex harmonic components in the cyclic frequency $\alpha$.

In summary, both Mask and Scale, as perturbation-based methods, only focus on the impact of the presence of features associated with the current class, which leads to a significant discrepancy from SHAP results. In contrast, SHEP can additionally capture the impact of the absence of features related to other classes, demonstrating good consistency with SHAP across various patch sizes, domains, and sample classes. It effectively approximates SHAP in a comprehensive manner.

\subsection{Comparison of Efficiency}
\label{subsec:Efficiency}

Patch-wise attribution and SHEP improve the efficiency of attribution calculation by reducing feature dimensions and computational complexity, respectively. The theoretical computational complexities of various methods are summarized in Table~\ref{tab:complexity}. Notably, due to the exponential complexity of enumeration-based SHAP, permutation-based SHAP is used as an approximation for SHAP in all experiments.

\begin{table}[htbp]
        \centering
        \caption{Computational Complexity of Different Methods. \label{tab:complexity}}
        \begin{threeparttable}
                \footnotesize
                \begin{tabular*}{\hsize}{@{\extracolsep{\fill}}ccccc}
\toprule[1pt]
Mask           & Scale             & SHEP            & \makecell[c]{SHAP\\(permutation)} & \makecell[c]{SHAP\\(enumeration)}                 \\\midrule[0.3pt]
$\mathcal{O}(d)$\tnote{a} & $\mathcal{O}(k_s\cdot d)$ & $\mathcal{O}(d\cdot n)$ & $\mathcal{O}(k_p\cdot d\cdot n)$                                           & $\mathcal{O}(2^d\cdot n)$ \\
\bottomrule[1pt]
                \end{tabular*}

                \begin{tablenotes}
                        \smallskip
                        \footnotesize
                        \item[a] Abbreviations: $d$ is the feature dimension, $k_s$ is the number of scale operations, $n$ is the number of background samples, and $k_p$ is the number of permutations. In experiments, $k_s=3$, $k_p=5$ and $n=5\times c$ where $c$ is the number of class.
                \end{tablenotes}
        \end{threeparttable}
\end{table}

\begin{figure}[htbp]
        \centering
        \includegraphics[width=8.5 cm]{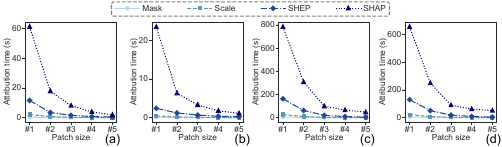}
        \caption{The computational times for a single sample across different attribution methods, patch sizes, and domains in the simulation dataset. (a)~Frequency domain. (b)~Envelope domain. (c)~Time-frequency domain. (d)~Cyclic-spectral domain.}
        \label{fig:4-4-SimuAttrTime}
\end{figure}

Theoretically, computational time depends on two primary factors: inference speed, determined by model complexity $\mathcal{O}(\mathcal{M})$ and hardware performance, and the number of required inferences. Since inference speed is beyond our research, we only focus on the latter. The computational time for a single sample across different methods, domains, and patch sizes is shown in Fig.~\ref{fig:4-4-SimuAttrTime} and Table~\ref{tab:4-4-SimuAttrTime}, which align with the theoretical complexities in Table~\ref{tab:complexity}. Regarding methods, Mask has the shortest computational time, followed by Scale, both of which have a complexity of $\mathcal{O}(d)$. SHEP and permutation-based SHAP exhibit significantly higher computational times due to their $\mathcal{O}(d n)$ complexity. However, SHEP reduces the number of permutations from $k_p$ to $1$ compared to SHAP, significantly lowering its computational time. Additionally, different domains and patch sizes affect the computational time primarily through the number of feature dimensions $d$. The 2D domain has a higher $d$ than the 1D domain and involves complex domain transforms $\mathcal{Z}$, leading to significantly higher computational times in the 2D domain.

\begin{table}[htbp]
        \centering
        \caption{The Computational Times for Each Sample across Different Attribution Methods, Patch Sizes, and Domains.}
        \label{tab:4-4-SimuAttrTime}
        \begin{threeparttable}
                \footnotesize
                \begin{tabular*}{\hsize}{@{\extracolsep{\fill}}ccccccc}
                        \toprule[1pt]
Domain                                                                       & Method                     & \#1\tnote{a} (s)    & \#2 (s)    & \#3 (s)   & \#4 (s)   & \#5 (s)   \\ \midrule[0.3pt]
\multirow{4}{*}{Freq}                                                   & \multicolumn{1}{c|}{Mask}  & 0.30   & 0.12   & 0.06  & 0.03  & 0.02  \\
                & \multicolumn{1}{c|}{Scale} & 2.08   & 0.70   & 0.26  & 0.10  & 0.05  \\
                & \multicolumn{1}{c|}{SHEP}  & 11.55  & 3.45   & 1.58  & 0.74  & 0.38  \\
                & \multicolumn{1}{c|}{SHAP}  & 61.26  & 18.21  & 17.71 & 17.81 & 18.17 \\\midrule[0.3pt]
\multirow{4}{*}{Env}                                                    & \multicolumn{1}{c|}{Mask}  & 0.09   & 0.04   & 0.03  & 0.02  & 0.01  \\
                & \multicolumn{1}{c|}{Scale} & 0.45   & 0.18   & 0.09  & 0.04  & 0.03  \\
                & \multicolumn{1}{c|}{SHEP}  & 2.40   & 1.21   & 0.66  & 0.35  & 0.23  \\
                & \multicolumn{1}{c|}{SHAP}  & 23.40  & 18.27  & 19.72 & 19.14 & 18.53 \\\midrule[0.3pt]
\multirow{4}{*}{TF}    & \multicolumn{1}{c|}{Mask}  & 5.04   & 2.07   & 0.72  & 0.37  & 0.16  \\
                & \multicolumn{1}{c|}{Scale} & 26.71  & 9.15   & 2.79  & 1.20  & 0.48  \\
                & \multicolumn{1}{c|}{SHEP}  & 162.48 & 61.57  & 19.79 & 9.27  & 3.64  \\
                & \multicolumn{1}{c|}{SHAP}  & 783.90 & 308.25 & 98.72 & 65.07 & 46.91 \\\midrule[0.3pt]
\multirow{4}{*}{CS} & \multicolumn{1}{c|}{Mask}  & 4.37   & 1.69   & 0.69  & 0.35  & 0.16  \\
                & \multicolumn{1}{c|}{Scale} & 21.87  & 7.14   & 2.60  & 1.14  & 0.44  \\
                & \multicolumn{1}{c|}{SHEP}  & 130.17 & 50.39  & 18.24 & 8.53  & 3.41  \\
                & \multicolumn{1}{c|}{SHAP}  & 656.88 & 250.31 & 90.25 & 61.73 & 50.62
\\ \bottomrule[1pt]
                \end{tabular*}
                \begin{tablenotes}
                        \footnotesize
                        \item[a] : the patch size level.
                \end{tablenotes}
        \end{threeparttable}
\end{table}

In summary, the proposed SHEP significantly reduces computational costs compared to the SHAP, although it cannot match the simpler methods like Mask and Scale. Both the domain and patch settings affect the feature dimensions $d$, and practical applications should carefully balance the need for detailed interpretability against computational efficiency to optimize performance.

\section{Experiment}

Under the same parameters, changing the dataset does not affect the efficiency of various attribution methods. Thus, this Section only discuss the interpretability performance.

\subsection{Open-source CWRU Bearing Dataset}

{\bfseries \itshape 1) Dataset Description:} The Case Western Reserve University (CWRU) bearing dataset is one of the most well-known open-source datasets for IFD, and we choose it to ensure the reproducibility. This experiment selects a load of 1 HP and a speed of 1800 rpm. The corresponding characteristic frequencies are shown in Table~\ref{tab:ExpCWRU-freq}. All samples are collected at a frequency of 12 kHz from the drive end, and they include four classes: healthy (H), inner race fault (I), rolling ball fault (B), and outer race fault (O), with fault sizes of 0.007 inches for each class. Each class contains 119 samples, and each sample has a length of 2000.
The experimental and training setups remain consistent with the simulation dataset. Ultimately, the end-to-end CNN model achieved a testing accuracy of 100\%.

\begin{table}[htbp]
        \centering
        \caption{The characteristic frequencies of two datasets.\label{tab:ExpCWRU-freq}}
        \begin{threeparttable}
                \footnotesize
                \begin{tabular*}{\hsize}{@{\extracolsep{\fill}}ccc|ccc}
                        \toprule[1pt]
                        \multicolumn{3}{c}{CWRU bearing dataset\tnote{a} (Hz)}                                              & \multicolumn{3}{c}{Helical gearbox dataset\tnote{b} (Hz)}       \\
                        \midrule[0.3pt]
                        $f_r$ & $f_{\mathrm{BPFI}}$ & $f_{\mathrm{BPFO}}$ & $f_1$      & $f_2$        & $f_{\mathrm{mesh}}$     \\
                        30   & 162.45                                & 107.55                                & 30   & 7.683 & 630 \\
                        \bottomrule[1pt]
                \end{tabular*}

                \begin{tablenotes}
                        \smallskip
                        \footnotesize
                        \item[a] $n_r=12$: the number of rolling elements.  $f_r$: the rotation frequency. $f_{\mathrm{BPFI}}$: the ball pass frequency with inner race. $f_{\mathrm{BPFO}}$: the ball pass frequency with outer race. 
                        \item[b] $f_1$: the drive gear frequency. $f_2$: the driven gear frequency. $f_{\mathrm{mesh}}$: the meshing frequency.
                \end{tablenotes}
        \end{threeparttable}
\end{table}

\begin{figure}[htbp]
        \centering
        \includegraphics[width=8.5 cm]{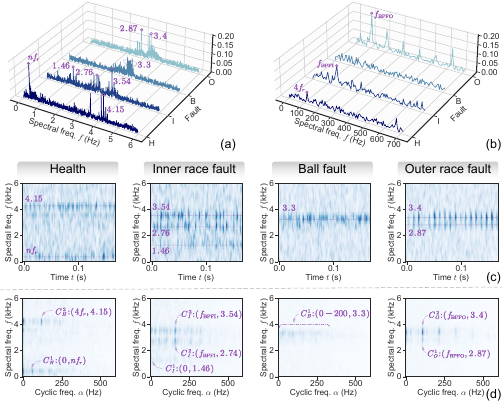}
        \caption{The domain representations of three classes in the CWRU dataset. (a)~Frequency domain. (b)~Envelope domain. (c)~Time-frequency domain. (d)~Cyclic-spectral domain.}
        \label{fig:Exp1Data}
\end{figure}

Unlike the simulation dataset with fully-known fault logic (ground truth), the CWRU dataset requires identifying the fault characteristics for each class firstly. For clarity, we denote modulated components as $C:(a,b)$, where $a$ (Hz) and $b$ (kHz) represent the carrier $f_c$ and modulation $f_m$ frequencies, respectively. These characteristics are highlighted in Fig.~\ref{fig:Exp1Data}. Specifically, the H-class sample includes a dominant constant frequency component $C_H^1\!:(0,nf_r)$ and a modulated component $C_H^2\!:(4f_r,4.15)$. The I-class sample includes $C_{I}^{1}\!:(0,1.46)$, $C_{I}^{2}\!:(f_{\mathrm{BPFI}},2.74)$, and $C_{I}^{3}\!:(f_{\mathrm{BPFI}},3.54)$. The B-class sample contains $C_{B}^{1}\!:(0-200,3.3)$. The O-class sample contains $C_{O}^{1}\!:(f_{\mathrm{BPFO}},2.87)$ and $C_{O}^{2}\!:(f_{\mathrm{BPFO}},3.4)$.

\begin{figure}[htbp]
        \centering
        \includegraphics[width=8.5 cm]{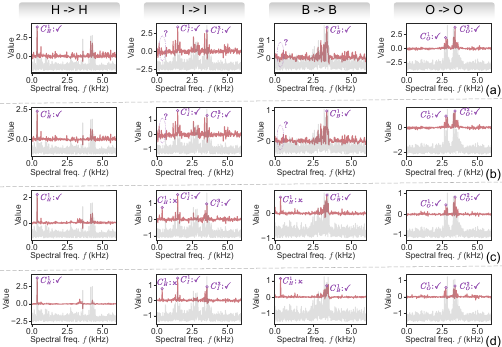}
        \caption{The frequency domain attribution results for the corresponding prediction classes of different class samples in the CWRU dataset under four attribution methods, with Patch \#1. (a)~Mask. (b)~Scale. (c)~SHEP. (d)~SHAP.}
        \label{fig:Exp1AttrFreq}
\end{figure}

{\bfseries \itshape 2) Interpretability Comparison:} The frequency attribution results of various methods  in the CWRU dataset are presented in Fig.~\ref{fig:Exp1AttrFreq}. For the H-class, $C_H^1$ serves as a key frequency, with its contribution due to its presence (i.e., $C_H^1\!:\!\checkmark$) in H-class samples accurately identified by both Mask and Scale methods. However, these methods fail to account for the contribution arising from its absence (i.e., $C_H^1\!:\!\usym{2613}\,$) in other class samples. In contrast, both SHEP and SHAP effectively capture the contribution of $C_H^1\!:\!\usym{2613}\,$ on the I and B class samples, demonstrating their superior interpretability.

\begin{figure}[htbp]
        \centering
        \includegraphics[width=8.5 cm]{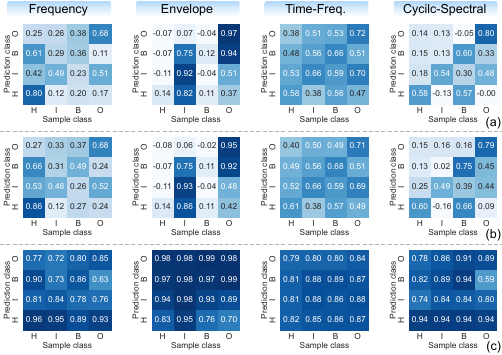}
        \caption{The cosine similarity matrices under different domains and attribution methods in the CWRU dataset, with Patch \#1. (a)~Mask. (b)~Scale. (c)~SHEP.}
        \label{fig:Exp1SimuMatrix}
\end{figure}

The cosine similarities for different domains and methods with Patch \#1 in the CWRU dataset are shown in Fig.~\ref{fig:Exp1SimuMatrix}. SHEP demonstrates very high cosine similarities, with most cases exceeding 0.8, which is significantly higher than those of the Mask and Scale methods. Notably, in the envelope domain, Mask and Scale methods perform poorly for H and B class samples. This is due to the absence of significant cyclic features in the H and B classes, such as $f_{\mathrm{BPFI}}$ in I-class or $f_{\mathrm{BPFO}}$ in O-class, making attribution more challenging. Despite this, SHAP maintains a very high similarity, showcasing its robustness.

\begin{figure}[htbp]
        \centering
        \includegraphics[width=8.5 cm]{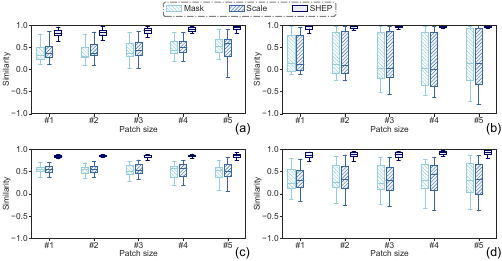}
        \caption{The statistic result of cosine similarity under different attribution methods and domains in the CWRU dataset. (a)~Frequency domain. (b)~Envelope domain. (c)~Time-frequency domain. (d)~Cyclic-spectral domain.}
        \label{Exp1SimuStatistic}
\end{figure}

The overall statistical result of cosine similarities in the CWRU dataset are shown in Fig.~\ref{Exp1SimuStatistic}, which exhibit a similar trend to the simulation dataset.  Specifically, as the patch size increases, the granularity of the explanations becomes coarser, generally resulting in higher similarity values. Besides, the similarity variance for Mask and Scale also increases, particularly in the Env and CS domains, which contain complex information related to cyclic frequency $\alpha$.

In summary, the open-source CWRU dataset demonstrates that SHEP maintains consistency with SHAP across various domains and patch sizes, significantly outperforming Mask and Scale methods.

\subsection{Private Helical Gearbox Dataset}

\begin{figure}[htbp]
        \centering
        \includegraphics[width=8.5 cm]{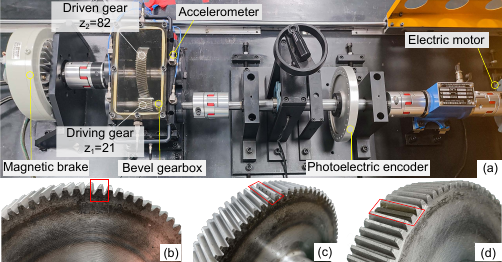}
        \caption{The experimental setup and fault types of the Helical gearbox. (a) The experimental setup. (b) W: Wear. (c) P: Pitting. (d) C: Tooth crack.}
        \label{fig:Exp2Rig}
\end{figure}

{\bfseries \itshape 1) Dataset Description:} This dataset was collected from a private helical gearbox to evaluate the effectiveness of SHEP in practical application. The experimental setup and fault types are depicted in Fig.~\ref{fig:Exp2Rig}. The motor was operated at 1800 rpm, with the characteristic frequencies listed in Table~\ref{tab:ExpCWRU-freq}. Data acquisition was performed at the drive shaft end-shield at a sampling rate of 5 kHz, encompassing four fault classes: healthy (H), wear (W), pitting (P), and crack (C). Each class comprised 76 samples, with each sample containing 2000 data points. The experimental and training procedures were consistent with those used for the simulation dataset. Ultimately, the end-to-end CNN model achieved a test accuracy of 100\%.

\begin{figure}[htbp]
        \centering
        \includegraphics[width=8.5 cm]{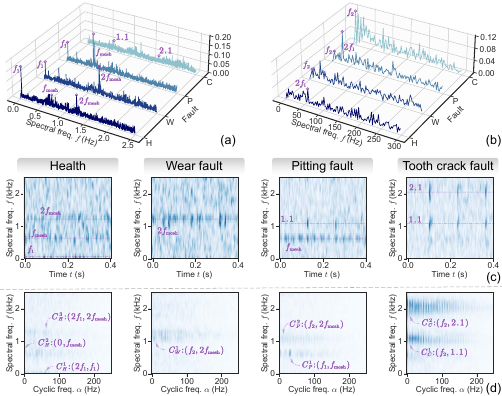}
        \caption{The domain representations of three classes in the helical gearbox dataset. (a)~Frequency domain. (b)~Envelope domain. (c)~Time-frequency domain. (d)~Cyclic-spectral domain.}
        \label{fig:Exp2Data}
\end{figure}

Similar to the CWRU dataset, ground-truth, i.e., the characteristics of each class, needs to be determined firstly. As shown in Fig.~\ref{fig:Exp2Data}, the H-class sample includes $C_H^1\!:(2f_1,f1)$, $C_H^2\!:(0,f_{\mathrm{mesh}})$, and $C_H^3\!:(2f_1,2f_{\mathrm{mesh}})$. The W-class sample contains $C_W^1\!:(f_2,2f_{\mathrm{mesh}})$. The P-class sample exhibits $C_{P}^{1}\!:(f_1,f_{\mathrm{mesh}})$, $C_{P}^{2}\!:(f_2,f_{\mathrm{mesh}})$. The C-class sample contains $C_{C}^{1}\!:(f_2,1.1)$, and $C_{C}^{2}\!:(f_2,2.1)$. 

\begin{figure}[htbp]
        \centering
        \includegraphics[width=8.5 cm]{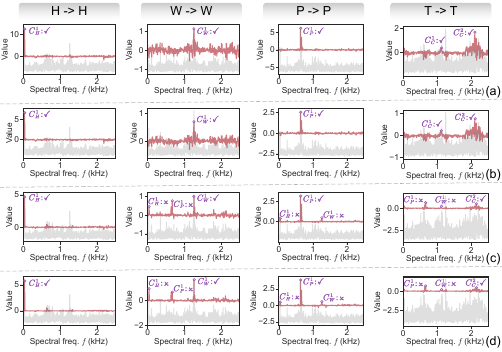}
        \caption{The frequency domain attribution results for the corresponding prediction classes of different class samples in the helical gearbox dataset under four attribution methods, with Patch \#1. (a)~Mask. (b)~Scale. (c)~SHEP. (d)~SHAP.}
        \label{fig:Exp2AttrFreq}
\end{figure}

{\bfseries \itshape 2) Interpretability Comparison:} The frequency attribution results of various methods applied to the helical gearbox dataset are presented in Fig.~\ref{fig:Exp2AttrFreq}.  Consistent with previous observations, Mask and SHAP account only for contributions arising from the presence of features within the current class (e.g., $C_H^1\!:\!\checkmark$ for H-Class, $C_W^1: \checkmark$ for W-Class). However, they fail to capture contributions resulting from the absence of features associated with other classes (e.g., $C_H^1\!:\!\usym{2613}\,$ for Classes W and P, $C_P^1\!:\!\usym{2613}\,$ for Classes W and C).

\begin{figure}[htbp]
        \centering
        \includegraphics[width=8.5 cm]{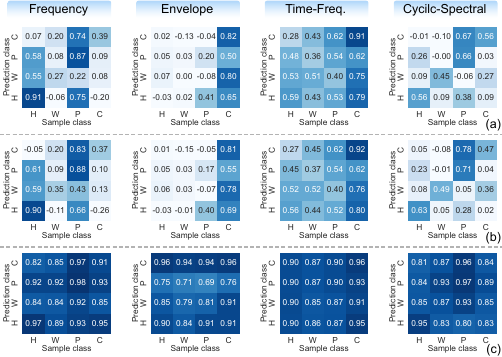}
        \caption{The cosine similarity matrices under different domains and attribution methods in the helical gearbox dataset, with Patch \#1. (a)~Mask. (b)~Scale. (c)~SHEP.}
        \label{fig:Exp2SimuMatrix}
\end{figure}

The cosine similarities for different domains and methods with Patch \#1 in the helical gearbox dataset are presented in Fig.~\ref{fig:Exp2SimuMatrix}. SHEP consistently outperforms Mask and Scale, whose similarities predominantly exceeding 0.8. In contrast, Mask and Scale exhibit notably lower similarities for Classes W and C in the Freq domain and for Classes H, W, and P in the Env domain. The reduced similarities in the Freq domain arises from their inability to capture the influence of $C_P^1\!:\!\usym{2613}\,$ on W-Class and $C_W^1\!:\!\usym{2613}\,$ on C-Class, as illustrated in Fig.~\ref{fig:Exp2AttrFreq}. The lower similarities in the Env domain can be attributed to the absence of distinct modulation frequencies in Classes H, W, and P, unlike the prominent modulated frequency $f_2$ seen in C-Class, as shown in Fig.~\ref{fig:Exp2Data}(b).

\begin{figure}[htbp]
        \centering
        \includegraphics[width=8.5 cm]{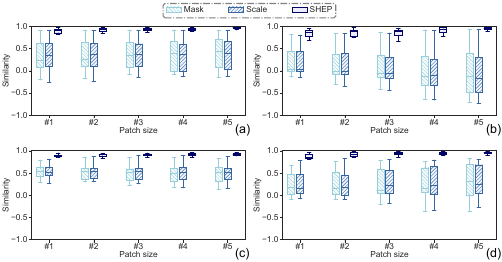}
        \caption{The statistic result of cosine similarity under different attribution methods and domains in the helical gearbox dataset. (a)~Frequency domain. (b)~Envelope domain. (c)~Time-frequency domain. (d)~Cyclic-spectral domain.}
        \label{fig:Exp2SimuStatistic}
\end{figure}

The overall statistical results of cosine similarities in the helical gearbox dataset are presented in Fig.~\ref{fig:Exp2SimuStatistic}. As the patch size increases, the explanations become coarser, leading to a remarkable improvement in SHEP's similarity to SHAP. Unlike the previous datasets, both Mask and SHEP exhibit high variance in similarities not only in the Env and CS domains, which include cyclic frequencies, but also in the Freq domain. This can be attributed to the higher diagnostic difficulty of the helical gearbox dataset, where multiple components share common carrier frequencies (e.g., $C_W^1$ and $C_P^2$) or modulation frequencies (e.g., $C_C^1$ and $C_C^2$).

In summary, the helical gearbox dataset poses a greater classification challenge, leading to lower similarity performance for Mask and Scale, as these methods only consider the contributions of feature presence. In contrast, SHEP accounts for both feature presence and absence, maintaining strong consistency with SHAP. This highlights the feasibility of approximating SHAP using SHEP, even in more complex diagnostic scenarios.

\section{Conclusion}
Due to the high dimensionality of vibration signals and the increased dimensions from domain transforms, SHAP in IFD encounters significant computational challenges. To address this, we propose patch-wise attribution and SHEP.
Key conclusions from extensive experiments are: 1): Patch size impacts explanation granularity linearly without affecting explanation correctness. 2): SHAP-\textit{Remove} effectively captures the contribution of feature presence, while SHAP-\textit{Add} identifies the impact of feature absence. Combining these, SHEP provides comprehensive explanations closely approximating SHAP. 3) SHEP demonstrates strong consistency with SHAP across diverse datasets, domains, patch sizes, and classes, validating its feasibility as an SHAP approximation. 4) Computational time and theoretical complexity align with expectations, with SHEP demonstrating significant efficiency advantages over SHAP.

In summary, the proposed methods enhance attribution efficiency and approximate SHAP effectively, offering feasibility for real-time interpretation in monitoring tasks. However, challenges remain: 1): Domain and patch size influence both explanations and computational cost, necessitating further research on task-specific parameter selection. 2): Interpretability evaluation remains difficult due to the absence of quantitative metrics and ground truth (e.g., fault logic) in real-world datasets.
We hope this work could inspire further research to address these challenge and improve the interpretability of IFD and other vibration-based tasks.

{\small
\bibliographystyle{ieee_fullname}
\bibliography{reference.bib}
}

\end{document}